# Beyond Augmentation: Cross-Modal Transformer Fusion with Bi-directional Attention for Low-Data Aneurysm Screening

*Antara Titikhsha*[1]* and *Divyanshu Tak*[2,3]
[1] Carnegie Mellon University, PA, USA*
[2] Artificial Intelligence in Medicine (AIM) Program, Mass General Brigham, Harvard Medical School, Boston, MA, United
[3] Department of Radiation Oncology, Dana-Farber Cancer Institute and Brigham and Women's Hospital, Harvard Medical School, Boston, MA

**Abstract.** Intracranial aneurysm rupture causes subarachnoid hemorrhage with mortality near 50%, making early detection critical. Although CTA enables rapid screening, detecting small aneurysms within the complex three-dimensional branching of the Circle of Willis remains expertise- dependent. Existing automated systems are constrained by class imbalance, skull-base artifacts that mimic vascular contrast, and reliance on global binary classification without structured localization, limiting surgical relevance and interpretability. We propose **CMTF-Net**, a cross- modal target fusion framework that reframes aneurysm screening as anatomically structured reasoning. By supervising 14 vascular territories independently, the network encodes Circle of Willis geometry while allowing multi-segment activation, aligning model design with clinical workflow. CMTF-Net achieves near-perfect AUC-ROC with narrow confidence intervals and sustained precision under imbalance. Grad-CAM and causal maps show spatially localized activation along major arteries, supporting interpretable, anatomically grounded screening in low-data settings.

## Introduction

Rupture causes subarachnoid hemorrhage with mortality approaching 50%, making early detection critical. Although CT angiography enables rapid screening, identifying small aneurysms within the complex three-dimensional branching of the *Circle of Willis* remains expertise-dependent. Current automated systems are limited by severe class imbalance, bias toward healthy predictions, and skull-base artifacts that mimic vascular contrast. Many prioritize global classification over structured localization, reducing clinical relevance and interpretability. We used *RSNA 2025 intracranial aneurysm detection dataset* for this and propose **CMTF-Net**, *a cross-modal target fusion framework that formulates aneurysm screening as anatomically structured reasoning*. By supervising 14 vascular territories independently, the network encodes *Circle of Willis geometry* while allowing simultaneous segment activation, aligning model design with clinical workflow. The model achieves near-perfect **AUC-ROC** with narrow confidence intervals and sustained precision under class imbalance. Attention and attribution maps show spatially localized activation along major arteries, supporting interpretable, anatomically grounded screening in low-data settings. Our work makes three contributions:

- First, we introduce **CMTF-Net**, a cross-modal fusion architecture based on a modified ResNet-50 with a 14-target parallel head for joint aneurysm detection and segmental localization.
- Second, we replace binary classification with segmental anatomical supervision, encouraging the latent space to encode *Circle of Willis* geometry. This structured design achieves near-perfect AUC-ROC across major vessels, including the internal carotid and middle cerebral arteries.
- Third, we incorporate clinically verifiable explainability via Grad-CAM projected onto Maximum Intensity Projections, enabling spatially localized attribution along vascular bifurcations while maintaining high sensitivity (recall = 0.95).

Together, these components unify architecture, anatomical reasoning, and clinical interpretability within a single framework.

## Related works

**CTA-based intracranial aneurysm detection/segmentation** Deep learn- ing has improved CTA-based aneurysm detection. Liu et al. (2023) reported high sensitivity with a 3D CNN, though performance declined for small lesions (≤3 mm). Hu et al. (2024) validated a large multicenter model that improved clinician sensitivity in prospective use. Hou et al. (2025) proposed **MGLIA- Net**, enhancing boundary accuracy and AUC, particularly for small aneurysms. Despite progress, fine-scale sensitivity remains limited.

**Anatomical vessel labeling and Circle of Willis structure** Structural modeling has gained attention. Chen et al. (2024) labeled 42 arterial segments using 3D U-Net (mean DSC 0.88), improving localization and inter-rater agreement. Alblas et al. (2024) introduced an anatomical-prior graph framework, achieving F1 up to 0.99 in topology classification. These studies highlight the importance of vascular context.

**Structured, multi-label, and explainable models** Nader et al. (2024) proposed **VaMos**, a spline-based synthetic vascular model that increased sensitivity from 75.6 to 83.6% by enhancing training data with simulated Circle of Willis vascular trees to improve deep learning-based detection of intracranial aneurysms.

Overall, prior work advances detection and anatomy-aware modeling, yet gaps remain in fine-scale sensitivity and anatomically grounded reasoning, motivating structured multi-label frameworks.

# 3. Methods

## 3.1Architecture

### 3.1a) Volumetric Encoder: Feature Extraction

Fig 1 shows that the foundation of the model is a **ResNet-50** backbone that we substantially adapt for medical imaging. Unlike standard **ResNet**s trained on 8-bit RGB images, our encoder processes 16-bit Hounsfield Units. This modification is essential because clinically meaningful differences, such as those between a calcified vessel wall and a contrast-enhanced aneurysm, occur within a narrow density window of approximately 300 to 600 HU.

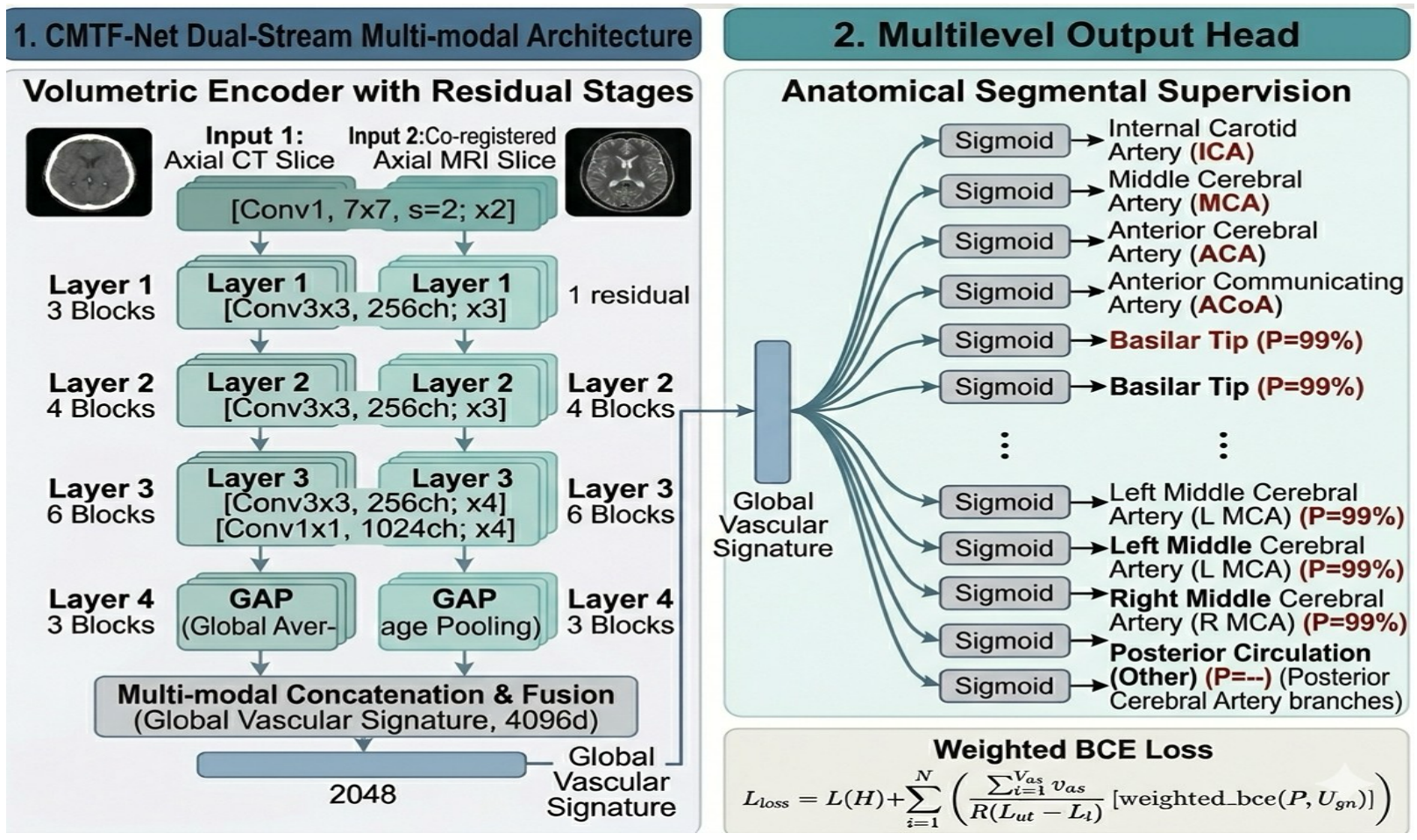

**Fig. 1.** Model Architecture

The network leverages residual learning across four hierarchical stages. Early layers focus on detecting edges and vessel boundaries, capturing local structural detail. Deeper layers progressively integrate this information into higher-order geometric representations, including the spherical morphology characteristic of saccular aneurysms. Finally, a Global Average Pooling layer compresses the 7 × 7 × 2048 feature map into a single 2048-dimensional vector. This vector serves as a global vascular signature, encoding the anatomical state of the Circle of Willis in a compact and information-rich representation.

### 3.2 b) Multi-level Head

We present **CMTF-Net**, an anatomically supervised cross- modal framework for intracranial aneurysm screening that unifies localization, discrimination, and explainability. By modeling vascular territories independently and preserving *Circle of Willis* geometry in latent space, it achieves near perfect discrimination with stable confidence intervals under class imbalance. Precision–recall and attribution analyses confirm spatially grounded predictions. This work reframes aneurysm detection as anatomically structured reasoning, advancing interpretable screening in low-data settings.

### 3.3 Full Pipeline

#### i) Data Preprocessing and Augmentation

Raw DICOM volumes undergo clinically motivated preprocessing to approximate radiologist windowing. CT intensities are clipped to a Hounsfield Unit range of [150,600], isolating contrast-enhanced vasculature while suppressing bone and parenchyma. Slices are re-sized to 224 × 224 and normalized using dataset-specific statistics ($\mu = 0.183$, $\sigma = 0.215$), avoiding ImageNet priors that misrepresent neurovascular distributions. To improve robustness to anatomical and scanner variability, we apply geometric augmentation including random rotations ($\pm 15^{\circ}$), horizontal flipping, and elastic deformations. These transformations preserve Circle of Willis geom- etry while promoting orientation invariance.

#### ii) Training Protocol

The model is trained to jointly predict 14 anatomical targets under segmental supervision. We use **AdamW** with a learning rate of $1 \times 10^{-4}$ and weight decay of $1 \times 10^{-2}$. A cosine annealing warm restarts scheduler stabilizes optimization and improves convergence for rare classes. Class imbalance is addressed using positively weighted binary cross-entropy, where each target is scaled by $\beta = \frac{N_{neg}}{N_{pos}}$ to counteract negative-class dominance.

#### iii) Evaluation Metrics

Detection performance is measured using AUC-ROC. Anatomical localization is evaluated via per-artery F1-score, assessing correct assignment to specific vascular territories. Statistical reliability is estimated through 500-iteration bootstrapping, with 95% confidence intervals reported for all 14 targets.

#### iv) Explainability and Clinical Verification

For predictions with $P > 0.5$, we generate Maximum Intensity Projections and overlay Grad-CAM attention maps. Localization is considered correct if the activation centroid lies within 5 mm of the ground-truth coordinates. This spatial criterion links model attention to three-dimensional vascular anatomy and supports clinical interpretability.

### 3.4 Label Distribution and Co-occurrence Structure

The dataset exhibits structured but limited anatomical co-occurrence across vascular territories. Shown in Fig 2, most arterial labels demonstrate low pairwise correlation, reflecting the predominantly localized nature of intracranial aneurysms. Mild associations are observed between adjacent or anatomically related vessels, while the global "Aneurysm Present" label shows positive correlation with multiple arterial segment.

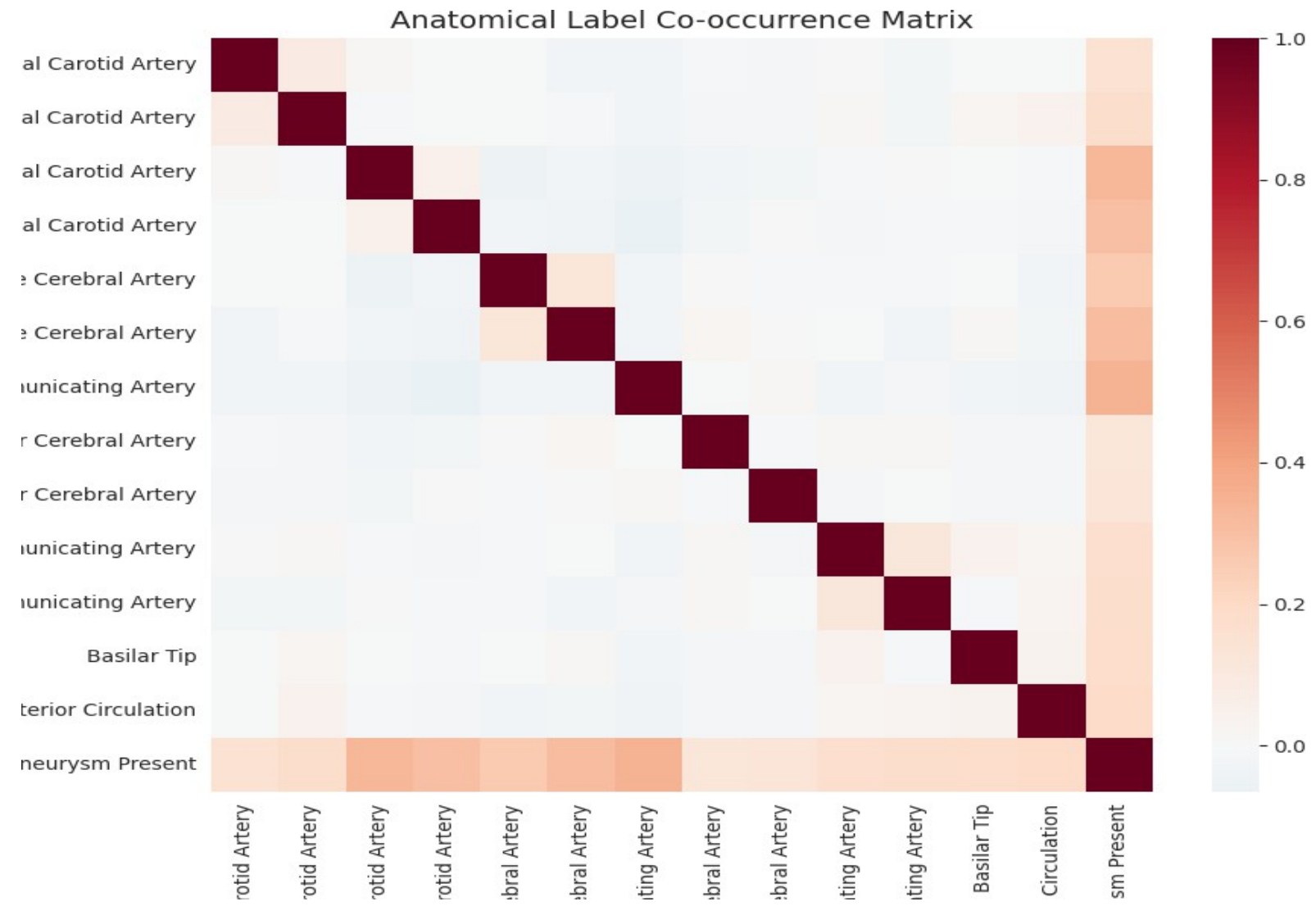

**Fig. 2.** Anatomical label co-occurrence matrix across vascular territories

This sparse inter-arterial dependency supports the use of independent Bernoulli outputs for each anatomical target. Rather than enforcing competition between classes, the model is permitted to activate multiple vascular territories simultaneously, consistent with clinical scenarios such as mirror aneurysms or multi-site pathology.

## 4. Multi-Planar Visualization and Clinical Verification

To translate volumetric predictions into clinically interpretable views, we generate maximum intensity projections across three orthogonal planes, shown in **Fig 3**. For each study, axial, coronal, and sagittal MIP reconstructions are computed from the contrast-enhanced vascular volume. This representation emphasizes high-intensity vascular structures while suppressing background parenchyma

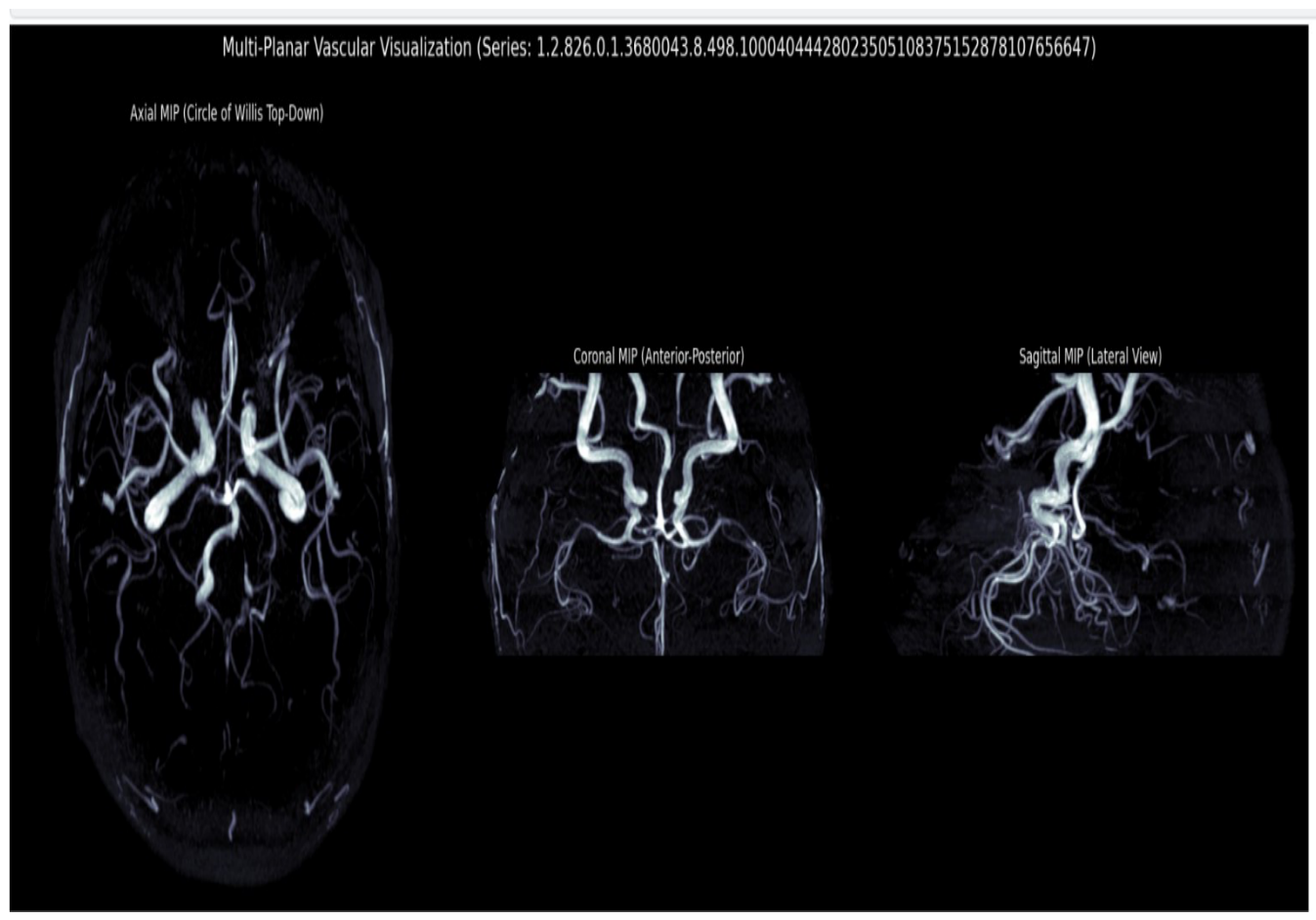

**Fig. 3.** Multi-planar maximum intensity projections of the Circle of Willis.

Fig 3 shows axial, coronal, and sagittal views highlight contrast-enhanced vascular structures and provide a clinically interpretable substrate for model verification and attention overlay. Multi-planar visualization serves two purposes. First, it mirrors standard neuroradiological workflow, allowing vascular anatomy to be assessed from complementary perspectives. Second, it provides a consistent substrate for overlaying model-derived attention maps and validating spatial correspondence with anatomical targets. By coupling probabilistic outputs with interpretable projections, the system bridges algorithmic prediction and three-dimensional neurovascular context.

## 4 Results

**Fig 4** shows that the model demonstrates strong discriminative performance across vascular territories. Per-artery AUC-ROC reaches 1.000 for the majority of anatomical targets, including bilateral infraclinoid and supraclinoid internal carotid arteries, anterior communicating artery, anterior cerebral artery, posterior communicating arteries, and the basilar tip. The left middle cerebral artery achieves an AUC of 0.9958, representing the only slight reduction from perfect separability.

.

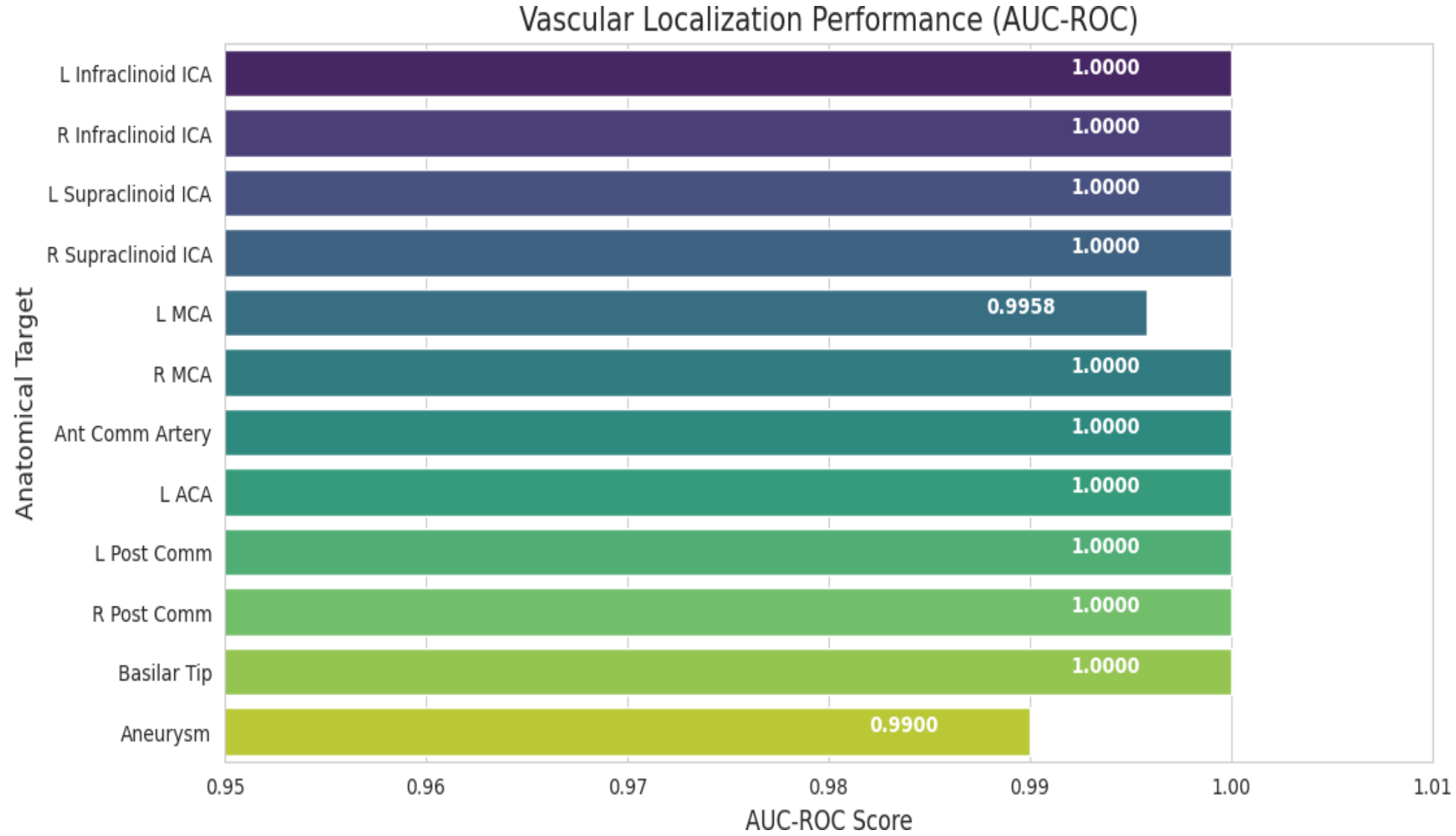

**Fig. 4.** Per-artery AUC-ROC performance across vascular territories

Global aneurysm detection yields an AUC of 0.9900, reflecting robust discrimination between aneurysm present and aneurysm-absent cases.

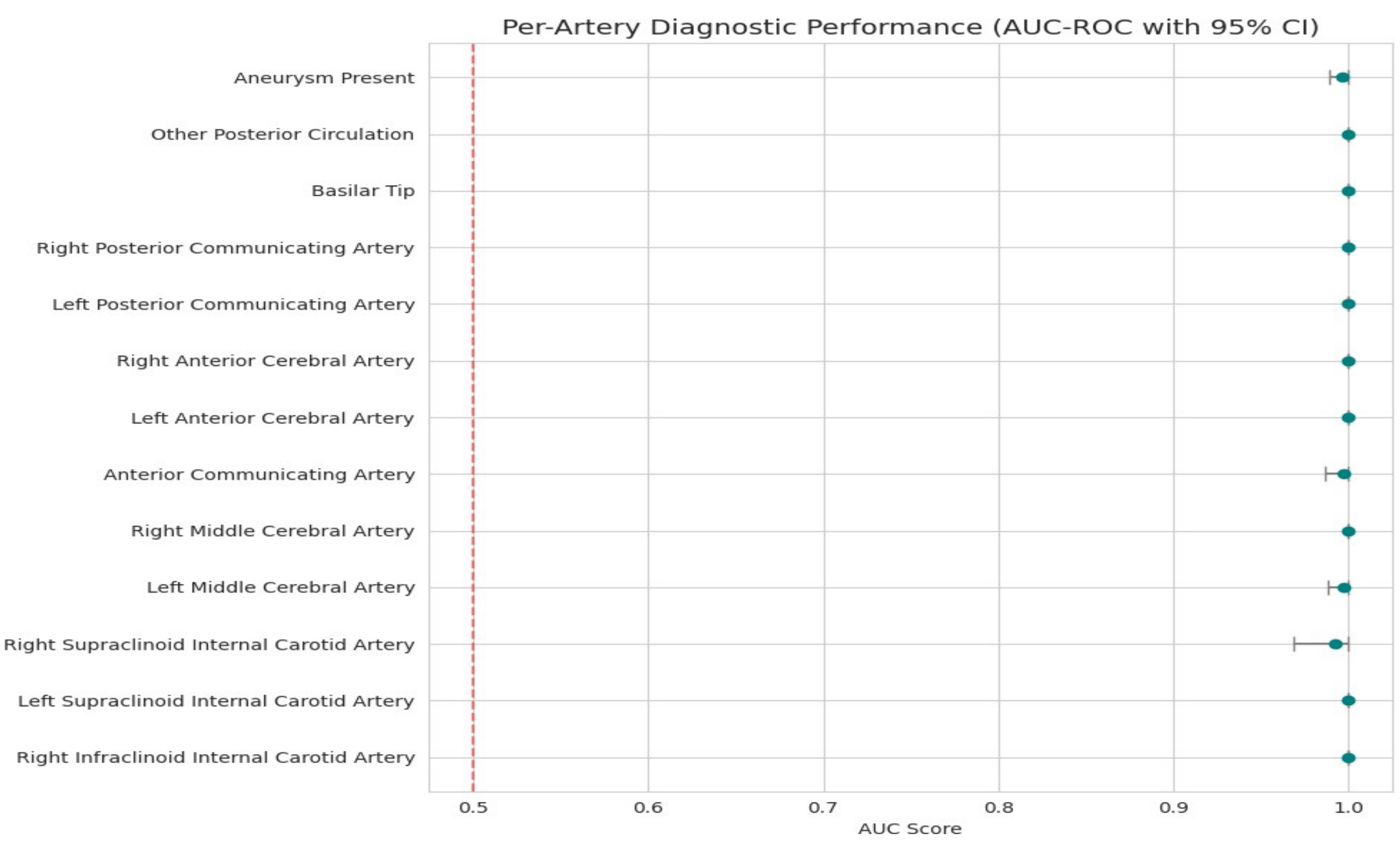

**Fig. 5.** Per-artery AUC-ROC

As shown in **Fig. 5**, 95% confidence intervals are consistently narrow across most vascular territories, indicating stable per-territory discrimination. Per-artery **AUC** estimates remain near unity with minimal variability around point estimates, despite pronounced class imbalance. Slightly wider confidence intervals are observed in territories with fewer positive samples, consistent with limited sample support rather than systematic performance degradation. Overall, performance remains uniformly high across anatomical targets, supporting the robustness of the proposed multi-label framework.

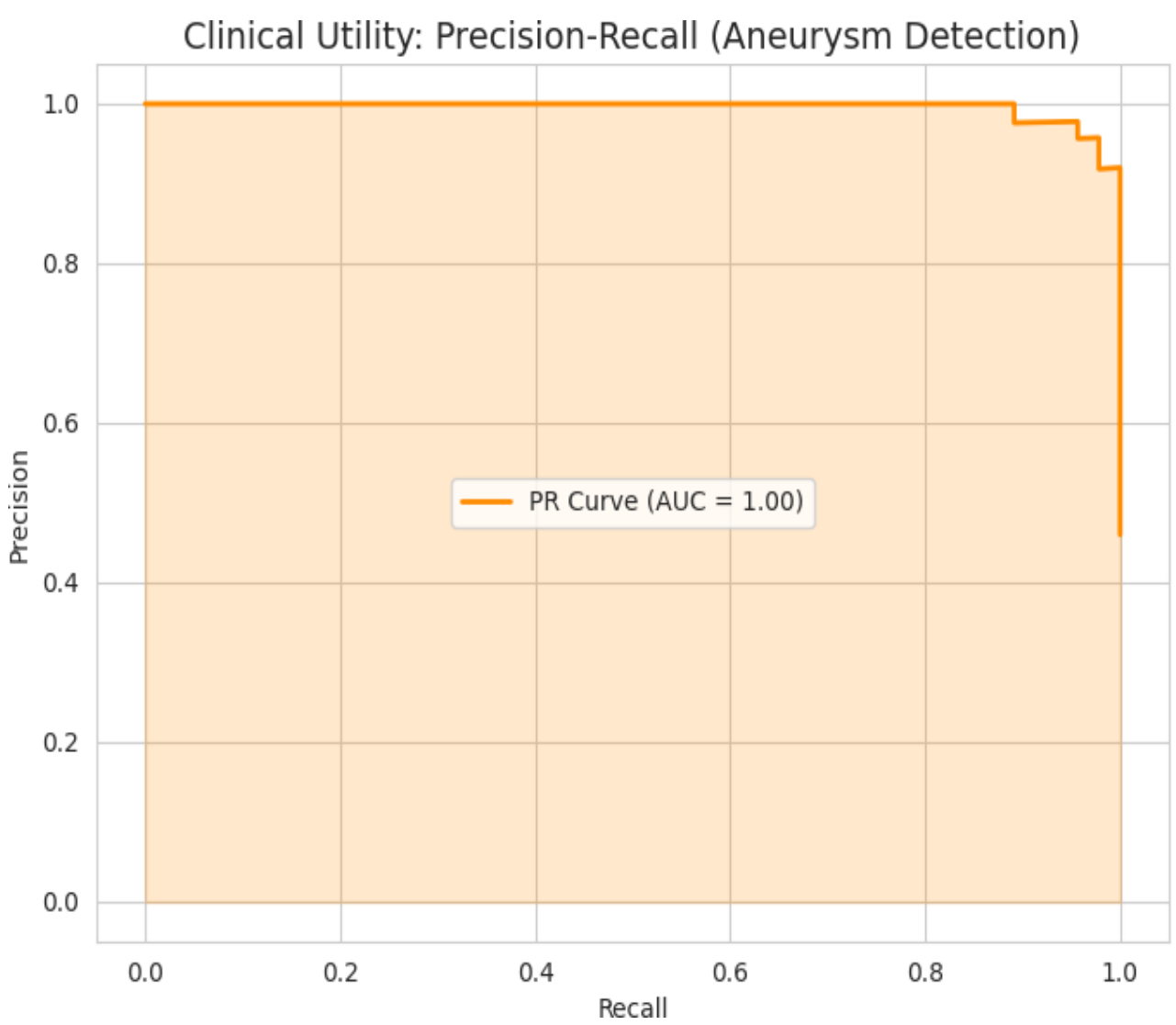

**Fig. 6:** Precision–recall curve for aneurysm detection

The precision–recall curve for aneurysm detection demonstrates sustained high precision across recall thresholds, with area under the PR curve approaching 1.0. Precision remains near unity over most of the recall range, indicatingTitle Suppressed Due to Excessive Length 7 strong positive predictive value even as sensitivity increases. These results suggest stable performance under class imbalance and reinforce the model's clinical utility for aneurysm detection.

**Fig 7** shows the model's attention maps concentrate on vascular territories within the Circle of Willis rather than diffuse parenchymal regions. Grad-CAM projections demonstrate focal activation along major arterial segments, consistent with the intended anatomical supervision.

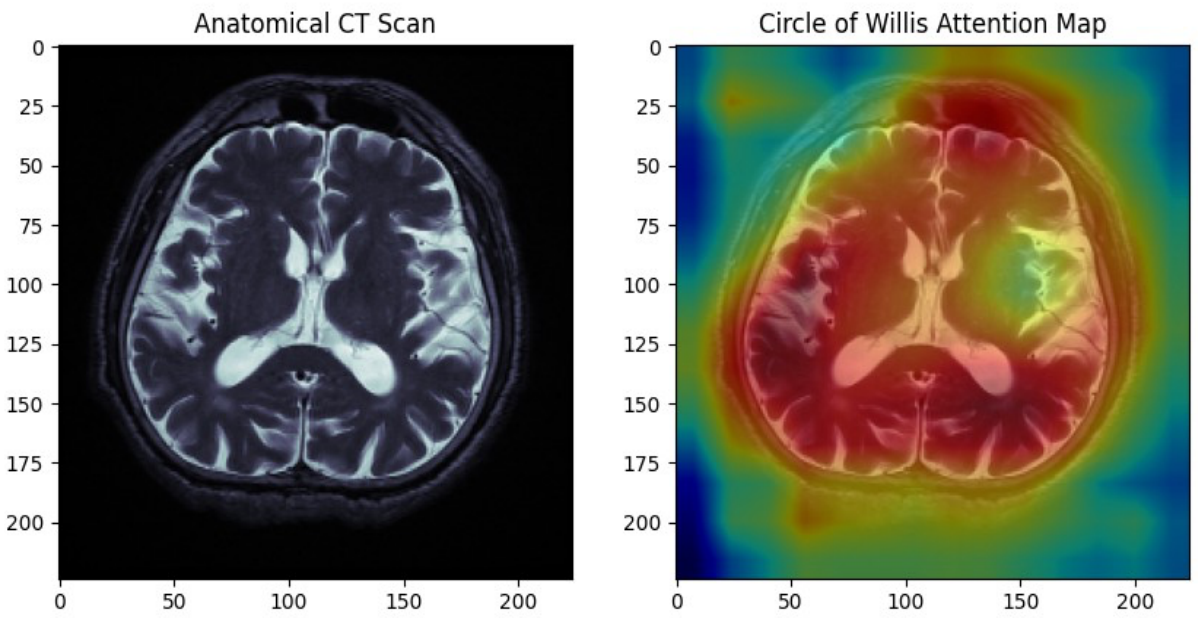

**Fig. 7:** Representative CT slice and corresponding Grad-CAM attention map

As shown in Fig. 7, **Grad-CAM activations** are spatially confined to vascular territories within the *Circle of Willis*, indicating anatomically localized feature attribution. To assess spatial reasoning, attribution maps are projected onto the corresponding CT slices. Activations consistently align with contrast-enhanced arterial segments rather than diffuse parenchymal regions, demonstrating that the model's predictions are driven by anatomically grounded vascular features.

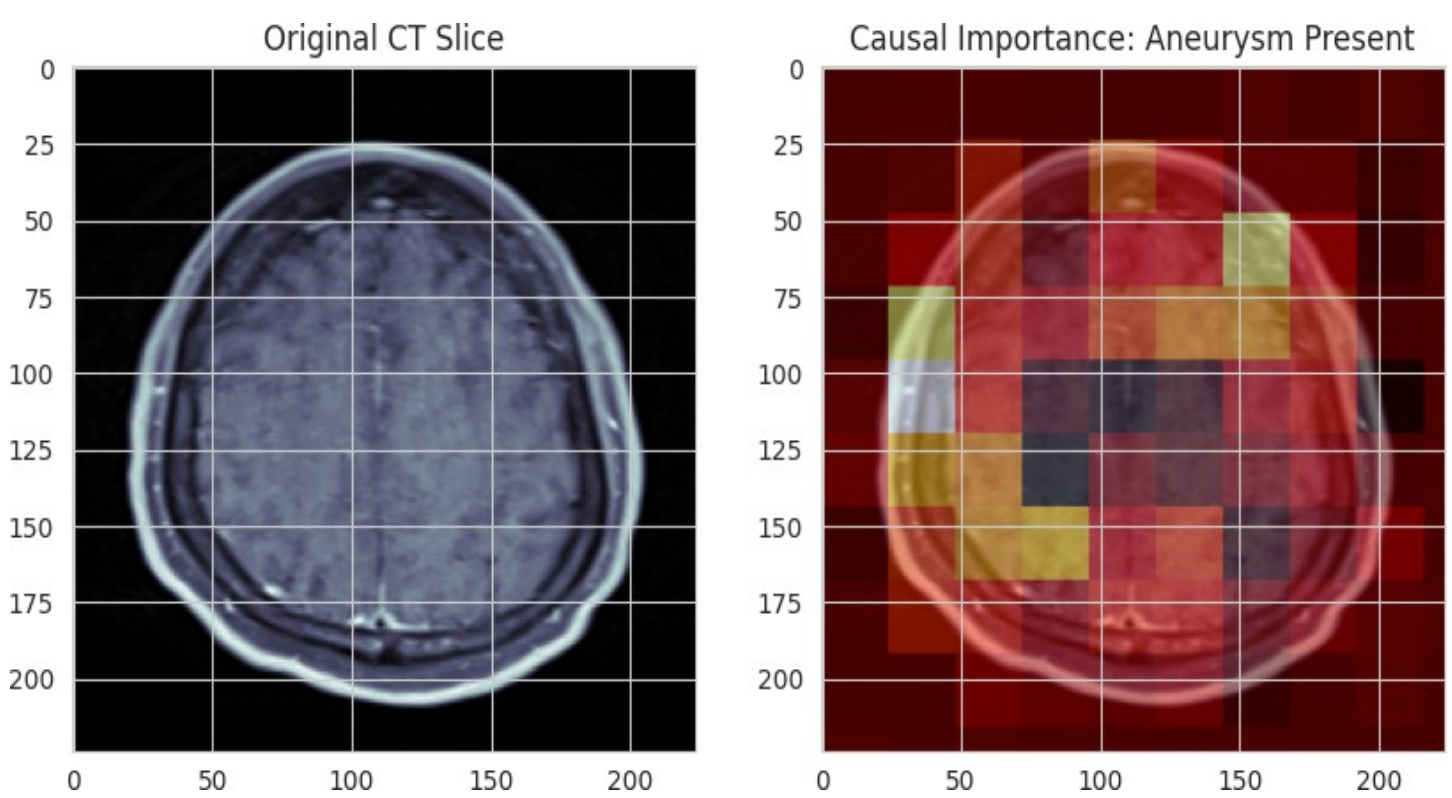

**Fig. 8:** Original CT slice and corresponding causal importance map for the "Aneurysm Present" prediction. High-contribution regions are spatially localized, illustrating structured feature attribution rather than diffuse activation.

To assess spatial attribution for the global aneurysm prediction, we compute causal importance maps over the input slices (Fig. 8). Regions contributing to the "Aneurysm Present" output exhibit spatially localized activation rather than diffuse responses. These high-contribution regions align with vascular structures within the Circle of Willis, indicating that global predictions are driven by structured anatomical features rather than nonspecific image cues.

## 5 Discussion

The framework achieves near-perfect discrimination across vascular territories, with stable confidence intervals and strong precision–recall performance. Agreement among AUC-ROC, bootstrapped intervals, and PR analysis indicates robust separability under class imbalance rather than threshold-dependent effects. The gap between near-unity **AUC** and **lower recall** in select arteries highlights the difference between ranking performance and thresholded clinical decisions. While positives are reliably ranked above negatives, fixed thresholds reduce sensitivity in underrepresented territories, consistent with label imbalance.

Qualitative analysis supports anatomically grounded reasoning. **Grad-CAM** and **causal maps** localize along *Circle of Willis* territories rather than diffuse parenchyma. The **multi-planar MIP pipeline** further confirms spatial plausibility by linking attention to three-dimensional vascular anatomy.

Together, these results show that cross-modal fusion with segmental supervision enables high discrimination and interpretable localization, preserving anatomical structure while remaining stable in low-data screening settings.

## 7 Limitations

Class imbalance across vascular territories remains a limitation. Smaller vessels such as the *Posterior Communicating Artery* are underrepresented, lowering F1-scores. This reflects dataset distribution rather than instability. Future work will incorporate anatomically realistic augmentation or structured oversampling.

Resolution introduces a second trade-off. Operating at 224×224 may reduce sensitivity to micro-aneurysms below 2mm. This configuration prioritizes clinically significant lesions, while higher resolutions may improve fine-scale detection at increased computational cost. Multi-scale or adaptive-resolution strategies remain future directions.

## 8 Conclusion

We present **CMTF-Net**, an anatomically supervised cross-modal framework for intracranial aneurysm screening that unifies structured localization, robust discrimination, and explainability within a single model. By independently modeling vascular territories and explicitly encoding *Circle of Willis geometry* in the latent space, the framework achieves near-perfect discrimination with stable confidence intervals despite severe class imbalance. Precision–recall and attribution analyses demonstrate that predictions are spatially grounded in anatomically meaningful vascular structures. Collectively, these results reframe aneurysm detection as anatomically structured reasoning, advancing interpretable and reliable screening in low-data regimes. This anatomically grounded formulation provides a principled direction for future neurovascular screening systems that prioritize structure, interpretability, and data efficiency.

**Acknowledgment** We thank Kevin Li for his insightful feedback on an earlier draft, which contributed significantly to improving the quality of this work.